\documentclass[final,3p,times,authoryear]{elsarticle}

\usepackage{amssymb,amsmath,graphicx,microtype,booktabs,hyperref, glossaries, algorithm, algpseudocode, graphicx, subfig, multirow, stfloats}

\journal{European Journal of Operational Research}

\begin{document}

\begin{frontmatter}

\title{A Hierarchical Importance-Guided Multi-objective Evolutionary Framework for Deep Neural Network Pruning}

\author{Zak Khan}
\author { Azam Asilian Bidgoli\corref{cor1}}
\cortext[cor1]{Corresponding author}
\ead{aasilianbidgoli@wlu.ca}
\affiliation{organization={Department of Computer Science and Physics, Wilfrid Laurier University},
    city={Waterloo},
    postcode={N2L 3C5},
    state={Ontario},
    country={Canada}
}

\begin{abstract}
The optimization of over-parameterized deep neural networks represents a large-scale, high-dimensional, and strongly non-convex decision problem that challenges existing optimization frameworks.  Current evolutionary and gradient-based pruning methods often struggle to scale to such dimensionalities, as they rely on flat search spaces, scalarized objectives, or repeated retraining, leading to premature convergence and prohibitive computational cost. This paper introduces a hierarchical importance-guided evolutionary  framework that reformulates convolutional network pruning as a tractable large-scale multi-objective optimization problem. In the first phase, a continuous evolutionary search performs coarse exploration of weight-wise pruning thresholds to shrink the search space and identify promising regions of the Pareto set. The second phase applies a fine-grained binary evolutionary optimization constrained to the surviving weights, where importance-aware sampling and adaptive variation operators refine local search  in the sparse region of the Pareto set. This hierarchical design combines global exploration and localized exploitation to achieve a well-distributed Pareto set of of networks balancing compactness and accuracy. Empirical results on CIFAR-10 and CIFAR-100 using ResNet-56 and ResNet-110 confirm the method’s effectiveness compared to existing evolutionary approaches: pruning achieves up to 51.9 \% and 38.9 \% parameter reductions with almost no accuracy  loss compared to state-of-the-art evolutionary DNN pruning methods.   The proposed method contributes a scalable evolutionary approach for solving very-large-scale multi-objective optimization problems, offering a general paradigm extendable to other domains where the decision space is exponentially large, objective functions are conflicting, and efficient trade-off discovery is essential.
\end{abstract}


\end{frontmatter}


\section{Introduction}

Deep Neural Networks (DNNs) have revolutionized artificial intelligence (AI) by enabling models to automatically learn complex patterns from large-scale data. Convolutional Neural Networks (CNNs), in particular, have been pivotal in this progress, achieving state-of-the-art performance in image classification, object detection, and segmentation tasks~\cite{Chung2024WIEA, Aras2025RobustCNNOptimization}. Their hierarchical structure allows progressive feature extraction, from low-level edges to high-level semantic concepts, without manual feature engineering~\cite{Chung2024WIEA}. However, as CNN architectures have become deeper to achieve higher accuracy, they have also grown substantially in computational and memory requirements. Modern CNNs can contain millions of parameters and incur significant computational overhead per inference~\cite{Lian2024MOCompressionCNN}. This over-parameterization poses a major challenge for optimizing large-scale CNNs and deploying them on resource-constrained systems, where energy efficiency and real-time inference are crucial~\cite{Louati2024BiLevelPruning,Chung2024WIEA, Aras2025RobustCNNOptimization}.

This growing computational burden highlights a central problem in deep learning: how to maintain predictive accuracy while substantially reducing model size and complexity. Model compression, which removes redundant pieces of the network, offers a direct solution but introduces a difficult optimization dilemma. Over-pruning sacrifices accuracy, while under-pruning limits efficiency. Balancing these conflicting objectives transforms pruning into a large-scale multi-objective optimization (MOO) \cite{Gunantara2018ReviewMOO}, where model accuracy and sparsity must be optimized concurrently.


Network pruning is a model compression technique that removes redundant or less significant parameters from a neural network while aiming to preserve predictive accuracy~\cite{Han2015DeepCC, dong2019networkpruning}. In the context of CNNs, pruning can target individual weights, filters, or entire channels to reduce computational cost and memory usage without severely degrading performance~\cite{Luo2017ThiNet, Molchanov2016Taylor, he2017channelpruning}. However, determining which components to remove presents a challenging optimization problem.
From an optimization standpoint, pruning deep CNNs can be viewed as a large-scale combinatorial problem ~\cite{li2022eapruning}, where each weight or structural element is a binary decision variable, either retained or removed. The resulting search space is vast, discrete, and highly non-convex, making direct optimization computationally infeasible.


Evolutionary Computation (EC) provides a natural mechanism for addressing such large-scale, non-differentiable optimization problems~\cite{Gunantara2018ReviewMOO, Coello2007EMOBook}. By evolving a population of candidate pruning configurations, Evolutionary Algorithms (EAs) are capable of exploring diverse regions of the search space and identifying trade-offs between model sparsity and accuracy~\cite{Fernandes2020ESPruning, Poyatos2022EvoPruneDeepTL}. In the context of CNN pruning, EC has been widely adopted due to its suitability for multi-objective optimization and its ability to simultaneously explore trade-offs between conflicting objectives~\cite{Chung2024WIEA}. Traditional optimization methods typically rely on scalarized loss functions, making it difficult to preserve balanced solutions when objectives such as accuracy, sparsity, and speed are inherently conflicting. Several papers have shown that EC-based pruning methods can effectively balance multiple objectives, even in highly non-convex and high-dimensional search spaces~\cite{Fernandes2020ESPruning, Lian2024MOCompressionCNN}. However, directly applying Multi-Objective Evolutionary Algorithms (MOEAs) to full-scale CNNs remains computationally prohibitive due to the enormous dimensionality of the search space, which often involves millions of binary decision variables. This challenge motivates a hierarchical optimization strategy that first reduces the search space through coarse-grained analysis before performing fine-grained evolutionary exploration. Such an approach enables scalable and efficient CNN pruning while preserving model performance.


Large-scale optimization in DNNs requires balancing two critical dimensions: global exploration and local exploitation. Coarse-grained optimization methods, when applied effectively, can map the high-dimensional search space into a smaller, more tractable region that still contains high-quality solutions~\cite{Zhang2008MOEAD}. In contrast, fine-grained binary optimization enables precise exploration within that region, refining parameter selections at a detailed level, and generating more solutions in sparse region of the Pareto front. Integrating these two perspectives within a unified optimization framework provides both scalability and precision.

In this context, a two-phase optimization paradigm emerges as a natural solution. The first phase performs coarse global optimization, reducing the dimensionality of the problem by identifying and removing redundant parameters through continuous, threshold selection~\cite{reyhan2024pruning}. This stage defines the feasible region for the second phase, which performs binary local optimization over the remaining parameters. The second phase treats pruning as a discrete multi-objective problem, refining sparsity patterns within the reduced search space identified earlier. This hierarchical division allows computational resources to be concentrated on promising regions, essential in large-scale evolutionary search.
By framing CNN pruning as a hierarchical large-scale optimization task, this study bridges the gap between computational efficiency and model performance, establishing a scalable pathway for MOO in deep neural networks.

\textbf{Contributions:
}
This study presents a hierarchical optimization perspective for CNN pruning, grounded in large-scale multi-objective optimization principles. The key contributions are summarized as follows:
\begin{enumerate}
    \item \textbf{Hierarchical Large-Scale Optimization:} Conceptualize CNN pruning as a two-level optimization process that first performs coarse global exploration via threshold-guided selection followed by fine-grained binary refinement, ensuring scalability to networks with millions of parameters.
   
    \item \textbf{Well-Distributed   Pareto Front:} The coarse Pareto front obtained from Phase~1 reveals a sparse, under-explored region of high-quality trade-offs; Phase~2 conducts localized binary search within this region to fill and refine the Pareto landscape with more well-covered solutions.
    \item \textbf{Importance-Guided MOO Formulation:} Phase~2 formulates pruning as a binary multi-objective problem, leveraging importance-aware criteria to exploit the sparse regions defined by Phase~1 while balancing accuracy and model complexity.
    
    \item \textbf{Comprehensive Experimentation:} The proposed framework is evaluated on multiple deep CNN architectures, including ResNet-18, ResNet-50, ResNet-56, ResNet-101, ResNet-110, and ResNet-152, using CIFAR-10 and CIFAR-100 datasets, demonstrating robustness and allowing comparison across pruning benchmarks in terms of model sparsity and accuracy.
\end{enumerate}

The remainder of this paper is organized as follows. Section~2 introduces the necessary background on convolutional neural networks, multi-objective optimization, evolutionary algorithms, and large-scale optimization concepts that motivate the proposed framework. Section~3 reviews related work on DNN pruning. Section~4 presents the proposed hierarchical two-phase evolutionary pruning framework, detailing the continuous global pruning stage and the importance-guided binary refinement stage. Section~5 reports comprehensive experimental results  across multiple ResNet architectures. Section~6 discusses the empirical findings, limitations, and implications of the proposed approach. Finally, Section~7 concludes the paper.

\section{Background}

DNNs and evolutionary optimization form the basis for the proposed two-phase pruning framework. This section provides a concise overview of CNNs, MOO, MOEA such as NSGA-II and MOEA/D, and principles from large-scale optimization that motivate the hierarchical design of our method.

\subsection{Convolutional Neural Networks}

CNNs are widely used for image recognition and related visual tasks due to their ability to extract hierarchical spatial features. A convolutional layer applies a set of learnable kernels to an input feature map, and computing outputs ~\cite{LeCun1998Gradient, Krizhevsky2012ImageNet}. Parameter sharing across spatial locations allows CNNs to scale efficiently, but deep architectures such as ResNets still contain millions of parameters~\cite{He2016ResNet}.

These models are known to be heavily over-parameterized, creating redundancy that can be removed through pruning~\cite{Han2015DeepCC, liu2017networkslimming}. However, pruning decisions at the weight or filter level lead to high-dimensional combinatorial search spaces, requiring advanced optimization techniques for effective exploration~\cite{Molchanov2016Taylor, Li2016PruningFilters}.

\subsection{Multi-Objective Optimization}

Pruning naturally involves conflicting criteria, most notably preserving accuracy while reducing model size. This makes the problem a multi-objective optimization problem, typically defined as:
\[
\min_{\mathbf{x} \in \Omega} \; \big(f_1(\mathbf{x}),\, f_2(\mathbf{x}),\, \dots,\, f_m(\mathbf{x})\big),
\]
where $\mathbf{x}$ is a candidate solution and $f_i$ are conflicting objectives~\cite{Coello2007EMOBook, Gunantara2018ReviewMOO}. A solution $\mathbf{x}_1$ \emph{dominates} $\mathbf{x}_2$ if:
\[
f_i(\mathbf{x}_1) \le f_i(\mathbf{x}_2)\ \forall i 
\quad \text{and} \quad
\exists j \; \text{s.t.}\; f_j(\mathbf{x}_1) < f_j(\mathbf{x}_2).
\]
The non-dominated solutions form the Pareto front, describing all optimal trade-offs.

Evolutionary Multi-Objective Optimization (EMO) methods avoid scalarization and do not assume convexity or differentiability, making them suitable for the discrete and non-convex setting of CNN pruning~\cite{Fonseca1993MOGA, Coello2007EMOBook}.

\smallskip
\noindent\textbf{NSGA-II}. The Non-dominated Sorting Genetic Algorithm II (NSGA-II)~\cite{Deb2002NSGAII} is one of the most widely adopted evolutionary multi-objective optimization algorithms. It leverages an efficient non-dominated sorting procedure to rank candidate solutions and uses a crowding-distance metric to maintain population diversity along the Pareto front. Selection is performed through a crowding-comparison tournament, followed by crossover and mutation operators that introduce variation and promote exploration of the search space. 

By sustaining a well-spread set of high-quality solutions near the true Pareto set, NSGA-II is particularly effective for uncovering accuracy–sparsity trade-offs in CNN pruning, where preserving both model performance and structural compactness is crucial.

\smallskip
\noindent\textbf{MOEA/D}. The Multi-Objective Evolutionary Algorithm based on Decomposition (MOEA/D)~\cite{Zhang2008MOEAD} optimizes multi-objective problems by decomposing them into a set of scalar subproblems, each associated with a weight vector. A common formulation is
\[
g(\mathbf{x} \mid \boldsymbol{\lambda}) =
\max_{i} \; \lambda_i \, | f_i(\mathbf{x}) - z_i^\ast |,
\]
where $\boldsymbol{\lambda}$ controls the trade-offs among objectives and $z_i^\ast$ denotes the ideal point. Rather than relying on dominance-based ranking, MOEA/D evolves a population in which each individual corresponds to a specific subproblem, and information is shared among neighboring subproblems. This enables efficient exploration of distinct regions of the Pareto front and is particularly effective for many-objective problems or those with complex front geometries.

MOEA/D can also be applied to settings with \emph{binary decision variables}, such as pruning masks in CNN compression. In such cases, binary variation operators (e.g., bit-flip mutation, uniform or two-point crossover) are used to generate new candidate solutions, while each subproblem’s weight vector guides the search toward a particular accuracy--sparsity trade-off. Neighborhood interactions help propagate promising binary structures, making MOEA/D well suited for optimization tasks.

\subsection{Large-Scale Optimization}

Large-scale optimization focuses on problems with very high-dimensional decision spaces \cite{Hotegni2023SparseMTL}, where traditional evolutionary algorithms often perform poorly due to loss of diversity, slow convergence, and difficulty modeling interactions among many variables. Recent research highlights that effective methods typically reduce the search dimensionality through decomposition, variable grouping, or subspace-based search \cite{Yokoyama2023MOHyperparameterGA}, allowing algorithms to explore simplified representations before refining solutions in higher-dimensional spaces.

In the context of DNNs, similar goals arise when attempting to optimize or train extremely large models. Techniques such as knowledge distillation \cite{Hinton2015Distillation}, quantization \cite{Jacob2018Quantization}, and parameter-efficient methods like LoRA \cite{Hu2021LoRA} address model complexity by reducing the number of effective parameters or simplifying the representation space. Pruning aligns with this broader perspective by explicitly removing redundant parameters, filters, or channels, thereby reducing dimensionality and enabling more efficient optimization. However, many existing works result in a very sparse Pareto front due to the complexity of these real-world optimization problems.

\section{Related Work}

Research on compressing CNNs spans several well-established directions, including unstructured pruning, structured pruning, single-objective evolutionary methods, and multi-objective evolutionary optimization. Unstructured pruning removes individual weights to induce fine-grained sparsity, while structured pruning eliminates filters or channels to produce hardware-friendly networks. Evolutionary methods, both single- and multi-objective, leverage population-based search to explore pruning spaces that are difficult to optimize with purely heuristic or greedy criteria~\cite{Chen2019EvolutionaryNetArchitecture, Anwar2017StructuredPruningDCNN}. Together, these strands form the foundation for modern pruning frameworks and motivate the importance-guided two-phase evolutionary strategy proposed in this work.

\subsection{Unstructured Pruning}

Unstructured pruning removes individual weights and typically relies on magnitude-based criteria. Early work by Han et al.~\cite{Han2015DeepCC} introduced iterative magnitude pruning with retraining, demonstrating that high sparsity can be achieved without catastrophic accuracy loss. Despite its simplicity and effectiveness, such irregular sparsity often limits direct hardware acceleration.

More recent work has explored improved importance estimation and global pruning objectives for unstructured CNN pruning. SNOWS ~\cite{Lucas2024PreservingDR} introduces a one-shot pruning framework that preserves intermediate representations by optimizing a global reconstruction objective, outperforming layer-wise magnitude heuristics at high sparsity levels. Evolutionary approaches have also been applied to unstructured pruning. ~\cite{reyhan2024pruning} employ evolutionary search to jointly evaluate sparsity and accuracy, discovering redundant connections at both neuron and weight levels without relying on fixed pruning thresholds. Phase~1 of this thesis aligns with this class of methods by applying continuous, threshold-based pruning to establish a global sparsity baseline before discrete optimization.


\subsection{Structured Pruning}

Structured pruning removes entire filters, channels, or blocks,  yielding models with direct hardware efficiency benefits. Early methods focused on heuristic importance criteria. ~\cite{Li2016PruningFilters} prune filters with the smallest $\ell_1$-norms, while ~\cite{hu2016networktrimming} propose the APoZ criterion, ranking filters by activation sparsity. ~\cite{Luo2017ThiNet} perform channel pruning by reconstructing feature maps to minimize information loss, demonstrating that structured sparsity can be achieved without severe accuracy degradation.

Importance-based structured methods include the original Taylor expansion approach of ~\cite{Molchanov2016Taylor}, which estimates filter saliency via first-order sensitivity, and its later refinement~\cite{Molchanov2019Importance}. These criteria capture global sensitivity but are typically applied greedily layer by layer, potentially propagating suboptimal decisions.

More recent work has shifted toward global and coordinated structured pruning formulations. ~\cite{Louati2024BiLevelPruning} cast joint filter and channel pruning as a bi-level optimization problem, enabling coordinated pruning decisions across layers rather than independent greedy selection. Evolutionary optimization has also been increasingly adopted for structured pruning. ~\cite{Chung2024WIEA} propose a multi-objective evolutionary architectural pruning framework with weights inheritance, allowing structured pruning configurations to be optimized while significantly reducing retraining cost. Similarly, ~\cite{Lian2024MOCompressionCNN} apply multi-objective evolutionary optimization to CNN compression, jointly considering accuracy, parameter count, and computational complexity.

\subsection{Evolutionary-based Pruning }
\textbf{Single-Objective evolutionary optimization:
}Evolutionary pruning methods apply evolutionary strategies (ES), genetic algorithms (GAs), or mutation-driven heuristics to optimize a scalar objective combining accuracy and model size. EvoPruneDeepTL~\cite{Poyatos2022EvoPruneDeepTL} evolves filter-level pruning masks for transfer learning, identifying redundant filters while preserving downstream representation quality. Other ES-based methods~\cite{Fernandes2020ESPruning} evolve pruning configurations via mutation-driven architectural changes, demonstrating superior exploration of the pruning search space compared to greedy rule-based approaches.


While single-objective evolutionary pruning demonstrates strong exploratory capabilities, scalarizing accuracy and sparsity enforces a predetermined trade-off. This limits visibility into the full Pareto trade space and motivates multi-objective evolutionary approaches~\cite{Hotegni2023SparseMTL, Yokoyama2023MOHyperparameterGA}.

\textbf{Multi-Objective evolutionary optimization:
}
MOEAs, such as NSGA-II, optimize conflicting objectives simultaneously, producing a Pareto frontier that exposes trade-offs between accuracy and model complexity. This makes them a natural fit for pruning and architecture compression~\cite{Aras2025RobustCNNOptimization}.

A prominent example is \emph{Multi-objective Evolutionary Architectural Pruning}~\cite{Chung2024WIEA}, which encodes architectural pruning decisions and applies NSGA-II to jointly minimize error and network size. A key contribution is the use of \emph{weights inheritance}, reducing retraining cost by reusing parent weights across generations. More recently, \emph{EAPruning}~\cite{li2022eapruning} applies evolutionary pruning to both CNNs and vision transformers, integrating importance-guided criteria within a multi-objective evolutionary framework to improve efficiency and accuracy.

As evolutionary approaches scale to large networks, weight inheritance, hierarchical search-space decomposition, and importance-guided selection have become increasingly common~\cite{Lian2024MOCompressionCNN, Louati2024BiLevelPruning}. The  two-phase pruning framework developed in this work is closely aligned with these trends: Phase 1 establishes a global pruning baseline through continuous thresholding, while Phase 2 performs importance-aware evolutionary search in a discrete binary pruning space to explore the full accuracy–sparsity Pareto landscape.

\section{Proposed Method: Hierarchical Importance-Guided Multi-objective  Framework}

\begin{figure}[h!]
    \centering
    \includegraphics[width=\linewidth]{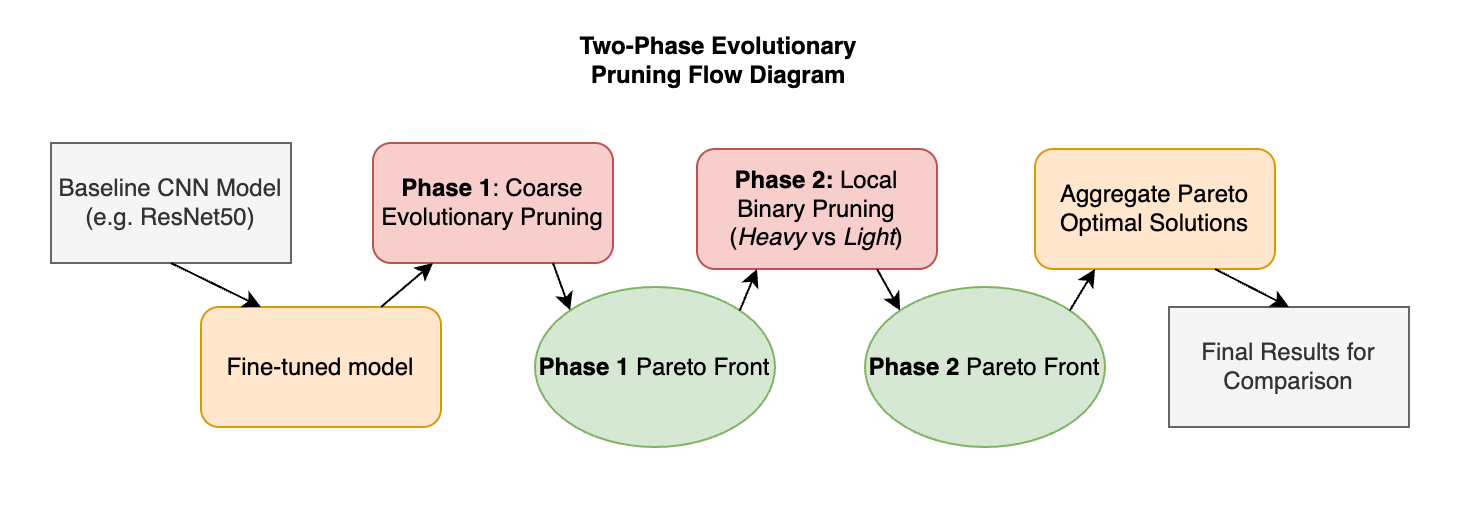}
    \caption{Two-Phase Evolutionary Pruning Flow Diagram. Phase~1 performs coarse-grained evolutionary pruning in continuous space, while Phase~2 refines the sparse region using binary importance-guided search to produce the final Pareto-optimal models.}
    \label{fig:flow1}
\end{figure}

The proposed pruning framework consists of two consecutive evolutionary MOO phases designed to progressively reduce model complexity while maintaining accuracy. As illustrated in Fig.~\ref{fig:flow1}, the process begins with a fine-tuned baseline CNN model (e.g., ResNet50), which undergoes a coarse-grained evolutionary search in Phase~1 to identify global pruning thresholds and generate the first Pareto front of sparsity–accuracy trade-offs. Phase~2 then performs a localized, binary-level evolutionary refinement within the \textit{sparse} region defined by Phase~1, using importance-guided initialization to search a diverse subspace. The final set of Pareto-optimal models represents a diverse range of compact, high-performing configurations. 

\subsection{ Objective Functions}

The proposed framework formulates CNN pruning as a hierarchical large-scale optimization problem with two conflicting objectives: minimizing model complexity and maximizing predictive accuracy. For a given pruning configuration $x$, these objectives are mathematically expressed in Eq.~\ref{eq:objectives}.

\begin{equation}
\begin{aligned}
f_1(x) &= \text{NonZeroWeights}(x), \\
f_2(x) &= 1 - \text{Accuracy}(x).
\end{aligned}
\label{eq:objectives}
\end{equation}

Here, $x$ denotes the decision variables associated with pruning: continuous thresholds in Phase~1 or binary masks in Phase~2. The first objective, $f_1(x)$, encourages network sparsity by minimizing the number of remaining non-zero weights, while the second objective, $f_2(x)$, preserves model performance by minimizing classification error. These two objectives are jointly optimized using MOEA, which constructs an approximate Pareto frontier representing the optimal trade-offs between sparsity and accuracy.

\subsection{Phase~1: Continuous Threshold-Based Pruning (Global Exploration)}

Phase 1 performs a coarse global optimization in the continuous space of pruning thresholds inspired by \cite{reyhan2024pruning}. Its objective is to explore the overall sparsity–accuracy landscape and identify promising regions for fine-grained refinement in Phase~2. Each individual in the population is defined by a pair of thresholds $(th_1, th_2)$ that specify the pruning interval, as expressed in Eq.~\ref{eq:prune_rule}.

\begin{equation}
w \in W, \quad 
w \gets 
\begin{cases}
0, & \text{if } th_1 \leq w \leq th_2,\\
w, & \text{otherwise},
\end{cases}
\label{eq:prune_rule}
\end{equation}

where $W$ denotes the complete set of learnable weights in the network. The threshold values determine the range of weight magnitudes that should be pruned by setting them to zero during evaluation.  The optimization process evolves the threshold pair $(th_1, th_2)$ using a MOEA which maintains a diverse population and balances both sparsity and accuracy through Pareto dominance and crowding distance mechanisms. 

The resulting non-dominated set of solutions forms the Phase~1 Pareto front, represented in Eq.~\ref{eq:pareto1}.

\begin{equation}
\mathcal{P}_1 = \{ (th_1^i, th_2^i) \,|\, i = 1,\dots,N \},
\label{eq:pareto1}
\end{equation}

This Pareto front $\mathcal{P}_1$ spans a spectrum of models, from highly accurate dense configurations to extremely sparse ones and defines the sparse Pareto region, which serves as the bounded feasible domain for Phase~2’s local binary refinement.

The entire Phase~1 process follows a standard multi-objective evolutionary optimization pipeline. The initial population is generated by randomly sampling real-valued threshold pairs $(th_1, th_2)$, enabling diverse coverage of the continuous pruning space. Each individual is evaluated by applying its threshold pair to the pretrained model and computing a bi-objective fitness: the number of remaining non-zero weights, measuring model sparsity, and the classification error on a validation dataset, measuring predictive performance. This process gradually refines a set of non-dominated threshold configurations that approximate the global sparsity–accuracy trade-off. The complete procedure for this phase is summarized in Algorithm~\ref{alg:phase1_pruning}, which details the initialization, evaluation, and evolutionary progression of threshold pairs.

\begin{algorithm}
\footnotesize
\caption{Phase~1: Coarse Global Search via Continuous Threshold Optimization}
\label{alg:phase1_pruning}
\begin{algorithmic}[1]
\Require Trained CNN model $M$, dataset $D$, population size $N$, number of generations $G$
\Ensure Pareto front $\mathcal{P}_1$ of optimal threshold pairs $(th_1, th_2)$
\State Initialize population $\mathcal{X}_0$ with random threshold pairs $(th_1, th_2)$
\For{each individual $(th_1, th_2) \in \mathcal{X}_0$}
    \State Prune $M$ by zeroing weights within $[th_1, th_2]$
    \State Evaluate objectives:
        \State $f_1(x)$ = number of remaining non-zero weights
        \State $f_2(x)$ = classification error on validation set $D$
\EndFor
\State Perform non-dominated sorting and compute crowding distances
\For{generation $t = 1$ to $G$}
    \State Select parents via tournament selection (based on Pareto rank and diversity)
    \State Generate offspring through simulated binary crossover and polynomial mutation
    \For{each offspring}
        \State Apply pruning using offspring’s $(th_1, th_2)$
        \State Evaluate $f_1$, $f_2$
    \EndFor
    \State Merge populations and update non-dominated set
\EndFor
\State \Return Pareto front $\mathcal{P}_1$ representing optimal sparsity–accuracy trade-offs
\end{algorithmic}
\end{algorithm}

\subsection{Phase~2: Importance-Guided Binary Pruning (Local Refinement)}

Phase~2 performs fine-grained optimization within the sparse region of the Pareto set identified by Phase~1, aiming to further reduce network complexity while preserving predictive accuracy. This refinement is formulated as a binary   multi-objective optimization problem solved via a MOEA. Each candidate solution encodes a binary pruning mask over the remaining non-zero weights ($N_r$) in the determined sparse region of the Phase~1 base model Eq. ~\ref{eq:binary_mask}:
\begin{equation}
x = [x_1, x_2, \dots, x_{N_r}], \qquad x_i \in \{0,1\},
\label{eq:binary_mask}
\end{equation}
where $x_i = 1$ retains the $i$-th weight and $x_i = 0$ prunes it. The optimization simultaneously minimizes the number of active weights and the classification error, producing a refined Pareto front of sparsity–accuracy trade-offs.



\subsubsection{Importance-Guided Probabilistic Initialization}

The objective of Phase~2 initialization is to generate a high-quality population of candidate pruning masks that operate within a defined sparsity range. Rather than initializing individuals randomly over the entire pruning space, this work constrains the initial population to a local search space identified from Phase~1, which allows focus on regions that exhibit favorable accuracy–sparsity trade-offs.

\noindent\textbf{Local Search Corridor}: To define this corridor, two anchor solutions are selected from the best Phase~1 Pareto front $\mathcal{P}_1$: a heavy model $M_h$ defining the beginning of the spare region and a light model $M_l$ defining the end, and together confining the broad sparse area within these solutions. These anchors correspond to Pareto-optimal solutions and are selected manually based on their non-zero weight counts, representing the upper and lower bounds of the sparsity range of interest. The heavy model $M_h$ contains $N_h$ non-zero weights and serves as the least-pruned reference, while the light model $M_l$ contains $N_l$ non-zero weights and represents an aggressively pruned but still competitive configuration. Together, the interval $[N_l, N_h]$ bounds the number of retained parameters explored in Phase~2.

To uniformly cover this interval, it is divided into $B$ bins $\{\mathcal{B}_1, \dots, \mathcal{B}_B\}$, each corresponding to a target range of non-zero weight counts. Multiple individuals are initialized per bin to ensure broad coverage across the sparsity spectrum. This bin-wise sampling strategy constrains the MOEA search space to a corridor of effective trade-offs.

\noindent\textbf{Importance-Guided Binary Initialization}: 
Phase~2 initializes each candidate solution as a binary pruning vector, where each element $x_{i,j} \in \{0,1\}$ determines whether weight $w_{i,j}$ in layer $i$ is retained or pruned. 
Each binary decision is sampled stochastically according to
\begin{equation}
x_{i,j} \sim \mathrm{Bernoulli}\!\left(1 - P_\text{prune}(w_{i,j})\right),
\label{eq:bernoulli}
\end{equation}
yielding
\[
\mathbb{P}[x_{i,j}=1]=1-P_\text{prune}(w_{i,j}), \quad 
\mathbb{P}[x_{i,j}=0]=P_\text{prune}(w_{i,j})
\]
Here, $x_{i,j}=1$ indicates that the weight is preserved, while $x_{i,j}=0$ denotes pruning. 
The pruning probability $P_\text{prune}(w_{i,j})$ is designed to incorporate both weight importance and layer-wise structural constraints, and is  defined as:
\begin{equation}
P_\text{prune}(w_{i,j}) = \lambda_i \cdot \big(1 - s(w_{i,j})\big),
\label{eq:prob_prune}
\end{equation}
where $s(w_{i,j}) \in [0,1]$ represents the normalized importance of the weight within its layer, and $\lambda_i \in [0,1]$ controls the pruning intensity applied to layer $i$. 
This formulation ensures that weights with higher importance are less likely to be pruned, while allowing deeper or less critical layers to undergo stronger pruning pressure.

The normalized importance score for each weight $w_{i,j}$ is computed using its magnitude relative to other weights in the same layer:
\begin{equation}
s(w_{i,j}) = \frac{|w_{i,j}|}{\max_k |w_{i,k}|}, \qquad s(w_{i,j}) \in [0,1],
\label{eq:importance}
\end{equation}
where larger values of $s(w_{i,j})$ indicate greater relative contribution to the layer’s representational capacity.

To enforce architectural stability and prevent over-pruning, each layer $i$ is assigned a sparsity-aware scaling factor $\lambda_i$ defined as
\begin{equation}
\lambda_i =
\begin{cases}
0, & \text{if } i \in \{L_n\}, \\[4pt]
0, & \text{if } \text{sparsity}(i) < S, \\[4pt]
\rho_i, & \text{otherwise, with } \rho_i \in [\alpha_i, \beta_i],
\end{cases}
\label{eq:lambda}
\end{equation}
where $L_n$ denotes layers excluded from pruning, $\text{sparsity}(i) = 1 - \tfrac{\text{NonZero}(W_i)}{|W_i|}$, and $W_i$ represents the set of learnable weights in layer $i$. 
The interval $[\alpha_i, \beta_i]$ specifies the allowable pruning ratio range for each layer and is treated as a tunable hyperparameter.

Equation~\eqref{eq:lambda} ensures that early convolutional layers, as well as layers whose sparsity falls below the threshold $S$, remain unpruned. 
This design reflects the critical role of early layers in learning low-level features and avoids destabilizing the network by pruning already sparse layers.
Together, Eqs.~\eqref{eq:bernoulli}–\eqref{eq:lambda} define an importance-guided probabilistic initialization mechanism that preserves critical weights while introducing controlled stochasticity to promote population diversity. 
The overall procedure is summarized in Algorithm~\ref{alg:smartinit}.

\begin{algorithm}
\footnotesize
\caption{Probabilistic Initialization for Binary Pruning}
\label{alg:smartinit}
\begin{algorithmic}[1]
\Require Pareto front $\mathcal{P}_1$, population size $N$, number of bins $B$, sparsity threshold $S$
\Ensure Initialized population $\mathcal{X}_0$
\State Select $M_h, M_l \in \mathcal{P}_1$ and define pruning corridor $[W_l, W_h]$
\State Partition $[W_l, W_h]$ into bins $\{\mathcal{B}_1,\dots,\mathcal{B}_B\}$
\For{each $\mathcal{B}_i$}
  \For{$j=1$ to $N/B$}
    \State Sample target non-zero count $K \in \mathcal{B}_i$
    \State Initialize binary mask $x$ of length $W_h$
    \For{each layer $i$ in $M_h$}
      \If{$i \in \{L_n\}$ or $\text{sparsity}(i) < S$}
        \State $x_{i,*} \gets 1$ \Comment{No pruning allowed}
      \Else
        \State Sample $\rho_i \in [\alpha_i,\beta_i]$ and compute $\lambda_i$ using Eq.~\eqref{eq:lambda}
        \State Compute normalized importance $s(w_{i,*})$ via Eq.~\eqref{eq:importance}
        \State Sample $x_{i,*}$ from Bernoulli as in Eq.~\eqref{eq:bernoulli}
      \EndIf
    \EndFor
    \State Normalize $x$ to match approximately $K$ retained weights
    \State Append $x$ to initial population $\mathcal{X}_0$
  \EndFor
\EndFor
\State \Return $\mathcal{X}_0$
\end{algorithmic}
\end{algorithm}


\noindent\textbf{Evolutionary Optimization via MOEA:} 
After importance-guided initialization, the binary population $\mathcal{X}_0$ is evolved using a MOEA to optimize the two competing objectives defined in Eq.~\eqref{eq:objectives}. At each generation, the evaluation of an individual proceeds by applying its binary pruning mask to the base model $M_h$, where weights corresponding to zero-valued mask entries are set to zero and retained weights remain unchanged. The resulting masked model is then evaluated on the validation set to compute classification error.

The model complexity objective is computed as the total number of non-zero weights remaining after masking, which directly reflects the pruning ratio achieved by the individual. This value is measured consistently across all individuals with respect to the same base model $M_h$. Based on these two objectives, individuals are ranked using the selected MOEA specific criteria, and the evolutionary process proceeds as detailed in Algorithm~\ref{alg:phase2}.

Importantly, the proposed importance-guided initialization, mask-based evaluation procedure, and constraint handling are independent of the underlying MOEA and are applied identically in both cases.  The influence of importance-aware pruning is instead introduced during the initialization stage, where pruning likelihoods $P_\text{prune}(w_{i,j})$ defined in Eq.~\eqref{eq:prob_prune} bias the initial population toward retaining structurally important connections. This design allows the evolutionary process to begin from high-quality, sparsity-aware candidates, while preserving algorithm independence and enabling subsequent exploration through uniform mutation.


\begin{algorithm}
\footnotesize
\caption{Phase~2: Importance-Guided Binary Evolutionary Pruning}
\label{alg:phase2}
\begin{algorithmic}[1]
\Require Initialized population $\mathcal{X}_0$, base model $M_h$, number of generations $G$
\Ensure Final Pareto front $\mathcal{P}_2$
\For{$t = 1$ to $G$}
  \For{each $x \in \mathcal{X}_{t-1}$}
    \State Apply mask $x$ to $M_h$
    \State Evaluate $f_1(x)$ and $f_2(x)$ using Eq.~\eqref{eq:objectives}
  \EndFor
  \State Apply MOEA operators (selection, crossover, importance-biased mutation)
  \State Form next generation $\mathcal{X}_t$
\EndFor
\State Extract $\mathcal{P}_2$ from $\mathcal{X}_G$
\State \Return $\mathcal{P}_2$
\end{algorithmic}
\end{algorithm}

\begin{figure}
    \centering
    \includegraphics[width=0.70\columnwidth]{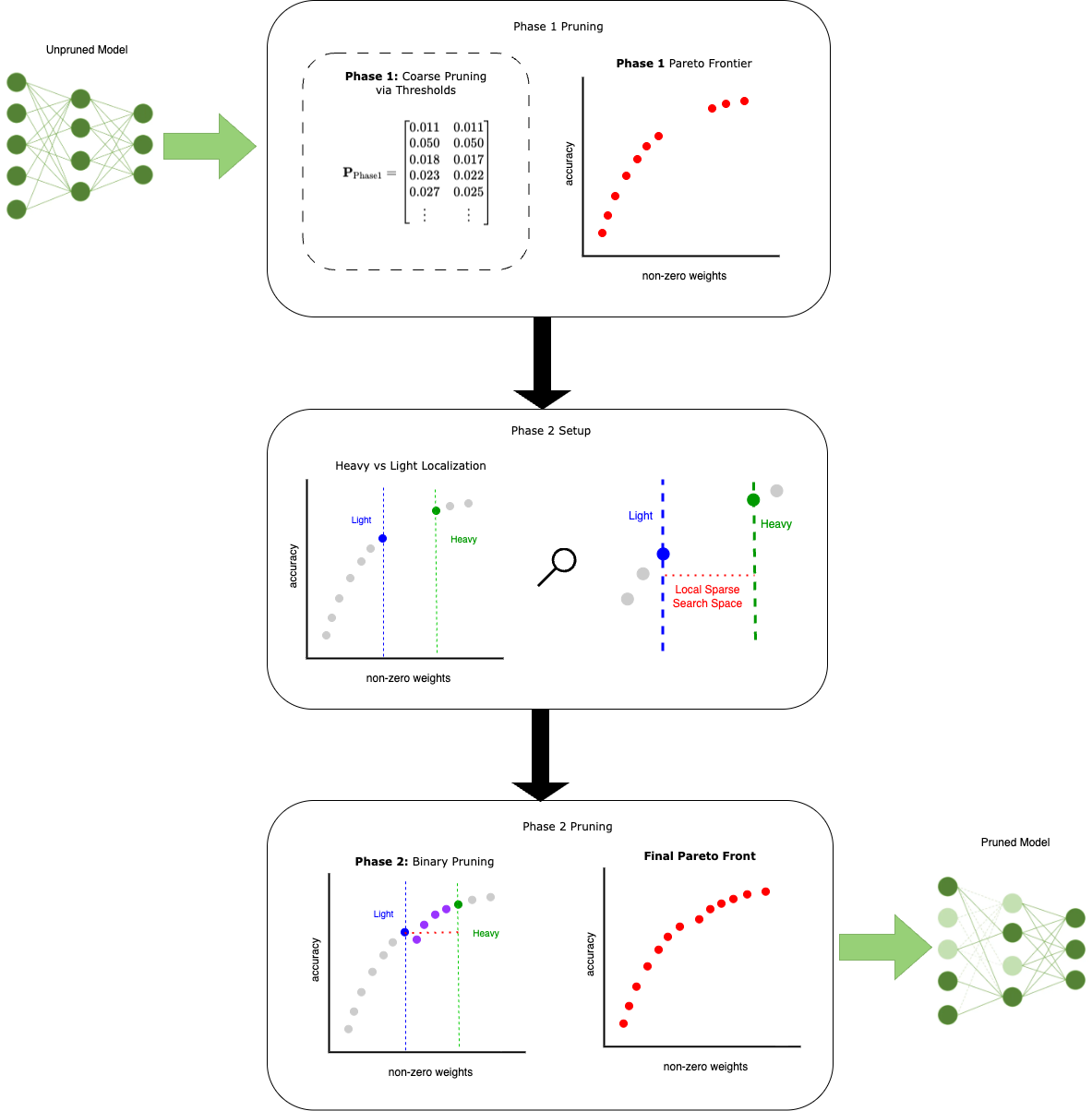}
    \caption{Overview of the two-phase pruning framework. Phase~1 constructs the Pareto front $\mathcal{P}_1$ via coarse threshold pruning, while Phase~2 performs localized binary refinement between heavy and light anchor models to produce the refined front $\mathcal{P}_2$.}
    \label{fig:flow4}
\end{figure}

The detailed optimization workflow is shown in Fig.~\ref{fig:flow4}, which depicts the transition from coarse-grained threshold pruning in Phase~1 to localized, binary pruning in Phase~2. The unpruned model starts off with a global threshold cutoff identifying sparsity and then refined locally in this region. In summary, Phase~1 evolves pruning thresholds through a standard evolutionary cycle of selection, crossover, and mutation, while Phase~2 extends this process with importance-guided initialization and binary mask evolution. Both phases iteratively evaluate individuals, fine-tune the resulting models, and extract Pareto-optimal solutions balancing sparsity and accuracy.




\section{Experiments and Results}

Before presenting the experimental results, this section describes the experimental setup used to evaluate the proposed two-phase pruning framework. The goal is to ensure transparency and reproducibility by detailing the computational environment, datasets, network architectures, training protocols, and evolutionary optimization settings employed in all experiments. This information provides the necessary context for interpreting the comparative results reported in the following sections.

\subsection{Experimental Setup}

This section outlines the configurations used to conduct all experiments in both pruning phases.  
It includes details on the computational environment, datasets, network architectures, fine-tuning protocols, and the hyperparameters used in the evolutionary optimization process.

\noindent\textbf{Hardware and Computational Resources.}  
All experiments were conducted on a workstation equipped with an Intel Xeon CPU (40 cores, 2.59\,GHz) and an NVIDIA RTX~A2000 GPU with 12\,GB of GDDR6 VRAM. Convolutional and gradient computations were performed on the GPU, while CPU threads managed data loading and evolutionary operations.



\noindent\textbf{Datasets and Preprocessing.}
Experiments were conducted on the CIFAR-10 and CIFAR-100 benchmark datasets~\cite{Krizhevsky2009CIFAR}, each containing $50{,}000$ training and $10{,}000$ test images of size $32\times32$. Images were resized to $224\times224$ and normalized using ImageNet statistics to match the input requirements of pretrained ResNet models~\cite{He2016ResNet}.

The CIFAR training set was partitioned into three disjoint subsets with distinct roles. The full training set was used to fine-tune the pretrained network. During each phase, a stratified optimization subset was constructed by sampling 50 images per class (500 total for CIFAR-10) or 10 images per class (1000 total for CIFAR-100) to enable computationally efficient evaluation of candidate solutions within MOEA. A separate validation subset of 1{,}000 or 2{,}000 images was randomly sampled from the remaining training data and used exclusively for model selection and generalization assessment. The original test set was reserved for final performance evaluation. Table~\ref{tab:dataset_summary} summarizes all subsets and their roles.

\begin{table}[H]
\centering
\small
\caption{Dataset composition and usage across all experimental stages.}
\label{tab:dataset_summary}
\resizebox{\columnwidth}{!}{%
\begin{tabular}{lcccc}
\toprule
\textbf{Dataset} & \textbf{Subset} & \textbf{Size} & \textbf{Purpose} & \textbf{Usage Phase} \\
\midrule
\multirow{4}{*}{CIFAR-10} 
& Training Set & 47{,}500 & Model fine-tuning & Pre-pruning \\
& Validation Set & 1{,}000 & Fitness evaluation & Phase 1 \& 2 \\
& Optimization Subset & 500 & Fast EA evaluation & Phase 1 \& 2 \\
& Test Set & 10{,}000 & Final model evaluation & Post-optimization \\
\midrule
\multirow{4}{*}{CIFAR-100} 
& Training Set & 47{,}000 & Model fine-tuning & Pre-pruning \\
& Validation Set & 2{,}000 & Fitness evaluation & Phase 1 \& 2 \\
& Optimization Subset & 1{,}000 & Fast EA evaluation & Phase 1 \& 2 \\
& Test Set & 10{,}000 & Final model evaluation & Post-optimization \\
\bottomrule
\end{tabular}%
}
\end{table}


\noindent\textbf{CNN Architectures.}  Six convolutional backbones were used throughout experimentation: ResNet-18, ResNet-50, ResNet-56, ResNet-101, ResNet-110, and ResNet-152 \cite{He2016ResNet}. The CIFAR-specific variants (ResNet-56 and ResNet-110) were used as benchmarks, as they are widely adopted reference architectures in CIFAR-based pruning and compression studies. Analyzing these ensured consistency while allowing comparison across varying network depths and parameter capacities. Table~\ref{tab:cnn_architectures} summarizes the architectural details.


\begin{table}[ht]
\centering
\caption{CNN architectures used in the experiments.}
\label{tab:cnn_architectures}
\begin{tabular}{lccc}
\toprule
\textbf{Model} & \textbf{Depth} & \textbf{Params (M)} & \textbf{Variant} \\
\midrule
ResNet-18  & 18  & 11.18 & ImageNet \\
ResNet-50  & 50  & 23.53 & ImageNet \\
ResNet-56  & 56  & 0.855 & CIFAR \\
ResNet-101 & 101 & 42.52 & ImageNet \\
ResNet-110 & 110 & 1.731 & CIFAR \\
ResNet-152 & 152 & 58.16 & ImageNet \\
\bottomrule
\end{tabular}
\end{table}

\noindent\textbf{Fine-Tuning Baseline.}  
Before pruning, all convolutional backbones were fine-tuned on the respective CIFAR datasets to establish strong baseline performance and stabilize weight distributions prior to evolutionary optimization. Each model was trained for 30~epochs using the Adam optimizer and a learning rate of $1\times10^{-4}$, with the cross-entropy loss function guiding the updates. See Table~\ref{tab:finetune_config}.  
  
\begin{table}[H]
\centering
\caption{Fine-tuning configuration and hyperparameter settings.}
\label{tab:finetune_config}
\begin{tabular}{ll}
\toprule
\textbf{Parameter} & \textbf{Setting} \\
\midrule
Optimizer & Adam \\
Learning rate & $1\times10^{-4}$ \\
Loss function & Cross-Entropy Loss \\
Epochs & 30 \\
Batch size & 32 \\
Scheduler & None (constant learning rate) \\

\bottomrule
\end{tabular}
\end{table}

\noindent\textbf{Evolutionary Hyperparameters.}  
Both pruning phases utilized population-based multi-objective optimization implemented in the \texttt{pymoo} framework \cite{Blank2020pymoo}.  
Phase~1 employed NSGA-II for continuous pruning-ratio optimization, while Phase~2 used both NSGA-II and MOEA/D for binary-mask refinement within the feasible region defined by the Phase~1 Pareto front.  
Each phase used a population of 50 and 50 generations to balance convergence and diversity. See Table~\ref{tab:evo_phase1} and Table~\ref{tab:evo_phase2_combined}. 

\begin{table}[H]
\centering
\caption{Evolutionary setup for Phase~1.}
\label{tab:evo_phase1}
\begin{tabular}{ll}
\toprule
\textbf{Parameter} & \textbf{Setting} \\
\midrule
Algorithm & \textbf{NSGA-II}\\
Population size / Generations & 50 / 50 \\
Sampling & Latin Hypercube Sampling (LHS) \\
Crossover & SBX ($p=0.9$, $\eta=15$) \\
Mutation & Polynomial ($p=0.2$, $\eta=20$) \\
Search space & Continuous $(r_i)$ \\
\bottomrule
\end{tabular}
\end{table}

\begin{table}[H]
\centering
\caption{Evolutionary setup for Phase~2.}
\label{tab:evo_phase2_combined}
\begin{tabular}{ll}
\toprule
\textbf{Parameter} & \textbf{Setting} \\
\midrule
Algorithm  & \textbf{NSGA-II} \\
Population size / Generations & 50 / 50 \\
Sampling & Importance-Guided Probabilistic Sampling \\
Crossover & Uniform ($p=0.9$) \\
Mutation & Bit-flip ($p=0.05$) \\
Search space & Binary pruning masks $(x_{i,j}\!\in\!\{0,1\})$ \\
\midrule
Algorithm  & \textbf{MOEA/D} \\
Population size / Generations & 50 / 50 \\
Sampling & Importance-Guided Probabilistic Sampling \\
Crossover & Uniform ($p=0.9$) \\
Mutation & Bit-flip ($p=0.05$) \\
Neighbors / Mating prob. & 15 / 0.9 \\
Search space & Binary pruning masks $(x_{i,j}\!\in\!\{0,1\})$ \\
\bottomrule
\end{tabular}
\end{table}

Phase~2 replaces the conventional binary-sampling with the importance-guided probabilistic sampling mechanism.  
Since MOEA/D is inherently a continuous MOEA, it was adapted to the binary pruning domain by optimizing continuous proxy variables and converting them into binary pruning masks via thresholding. Specifically, each candidate solution was mapped to a binary mask $x_{i,j}\!\in\!\{0,1\}$, where weights with values exceeding a predefined threshold were retained, and others were pruned.

\subsection{Baseline Performance}

Prior to applying pruning or evolutionary optimization, each ResNet baseline was fine-tuned on CIFAR-10 and CIFAR-100 to establish a stable and comparable starting point. The ResNet family was selected due to its modular architecture, and widespread use in pruning literature. 
Table~\ref{tab:baseline_errors} reports the resulting test error rates with all weights in the original network, which serve as baselines for evaluating performance changes in Phases~1 and~2 and for comparing pruning rates with other state-of-the-art methods.

\begin{table}[H]
\centering
\begin{tabular}{lcc}
\toprule
\textbf{Architecture} & \textbf{Test Error CIFAR-10} & \textbf{Test Error CIFAR-100} \\
\midrule
ResNet18   & 6.32  & 23.74 \\
ResNet50   & 6.03  & 22.08 \\
ResNet56   & 7.33  & 31.88 \\
ResNet101  & 5.34  & 20.76 \\
ResNet110  & 6.82  & 36.62 \\
ResNet152  & 5.17  & 20.56 \\
\bottomrule
\end{tabular}
\caption{Test error rates (\%) of baseline CNN architectures on CIFAR-10 and CIFAR-100.}
\label{tab:baseline_errors}
\end{table}

\subsection{Comprehensive Results Across Architectures}

Before presenting the results, we briefly define the evaluation metrics reported in Table~\ref{tab:merged_comparison} and Table~\ref{tab:merged_comparison_C100}. All results are averaged over 10 independent runs. Baseline accuracy refers to the test accuracy of the original unpruned model, and \# Weights denotes the total number of non-zero parameters (in millions) after preprocessing (to remove structurally redundant parameters specific to the CIFAR datasets). In Phase~1, the best solution (reported as non-zero weights and accuracy) corresponds to the highest-accuracy point on the continuous pruning Pareto front. The heavy anchor refers to the minimally pruned solution located near the dense end of the Pareto front, prioritizing accuracy with only slight parameter reduction. In contrast, the light anchor denotes a more aggressively pruned solution at the sparse end of the Pareto front, prioritizing model compression while tolerating larger accuracy degradation. These two anchors define the search interval explored in Phase~2. The reported difference (H-L) indicates the parameter span between these anchor solutions. Finally, Phase 1 HV denotes the hypervolume (HV) achieved after Phase~1 optimization.

Phase~2 results are reported on two well-known algorithms, NSGA-II and MOEA/D. \# Phase 2 Pareto solutions refer to the number of non-dominated solutions generated during Phase 2, while Phase 2 Pareto solutions dominating Phase 1  denote the subset of Phase~2 solutions that strictly dominate the light anchor solution from Phase~1 in the accuracy--sparsity objective space. Therefore, they remain non-dominated when merged with the Phase~1 Pareto front. Finally, HV measures the combined hypervolume of the Pareto front obtained after integrating both Phase~1 and Phase~2 solutions. An increase in this value compared with  Phase 1 HV equates to a more densely populated Pareto front, filling up sparse regions.

\noindent\textbf{CIFAR-10}. Table~\ref{tab:merged_comparison} summarizes the pruning performance of the proposed two-phase framework across six ResNet architectures on the CIFAR-10 dataset, averaged over ten independent runs. The corresponding Pareto fronts are visualized in Fig.~\ref{fig:PF_combined}. Across all models, the baseline networks achieve strong initial accuracies (92--95\%), and the preprocessing step reduces between 3\% and 10\% of redundant parameters without affecting predictive performance. This ensures that the subsequent pruning stages operate on compact and structurally valid models.
\begin{table*}[htbp]
\centering
\scriptsize
\caption{Pruning results for CIFAR-10. Baseline model size and accuracy, Phase 1 pruning outcomes (best, heavy, light anchor solutions), and Phase 2 multi-objective results (number of local and non-dominated Pareto solutions and the combined hypervolume). Results averaged over 10 runs.}
\label{tab:merged_comparison}
\resizebox{\textwidth}{!}{%
\begin{tabular}{lcccccc}
\toprule
\textbf{Metric} & \textbf{ResNet18} & \textbf{ResNet50} & \textbf{ResNet56} & \textbf{ResNet101} & \textbf{ResNet110} & \textbf{ResNet152} \\
\midrule
\multicolumn{7}{c}{\textit{Baseline (M)}} \\
\midrule
Baseline Weights (M)  & 11.18    & 23.53  & 0.855    & 42.52  & 1.731  & 58.16 \\
Baseline Accuracy (\%)            & 93.68    & 93.97 & 92.67 & 94.66 & 93.18 & 94.83 \\
\midrule
\multicolumn{7}{c}{\textit{Phase 1 Results}} \\
\midrule
Best Solution-Non-Zero Weights (M) & 6.970   & 12.02  & 0.506  & 17.14  & 1.575  & 22.62 \\
Best Solution Accuracy (\%)         & 93.14    & 93.49 & 92.18 & 93.64 & 92.85 & 93.35 \\
Heavy Anchor Weights (M)            & 4.767    & 9.684  & 0.483  & 16.15  & 1.473  & 21.83 \\
Heavy Accuracy (\%)                 & 93.10    & 92.98 & 91.97 & 92.82 & 92.14 & 92.94 \\
Light Anchor Weights (M)            & 3.812    & 6.342  & 0.387  & 13.96  & 1.036  & 17.84 \\
Light Accuracy (\%)                 & 91.67    & 88.40 & 88.84 & 87.93 & 88.51 & 86.51 \\
Difference (H–L, M)                 & 0.955    & 4.34  & 0.096 & 2.190  & 0.437  & 3.992 \\
Phase 1 HV                          & 0.813    & 0.807 & 0.824 & 0.818 & 0.806 & 0.821 \\
\midrule
\multicolumn{7}{c}{\textit{Phase 2 Results (NSGA-II)}} \\
\midrule
\# Phase 2 Pareto Solutions                     & 7.10    & 7.70  & 4.50  & 8.20  & 5.00  & 8.30 \\
\# Phase 2 Pareto Solutions Dominating Phase 1                            & 2.80   & 3.10  & 1.90  & 3.60  & 2.20  & 3.70 \\
Final HV (Phase 1 + Phase 2)   & 0.857    & 0.862 & 0.851 & 0.869 & 0.859 & 0.879 \\
\midrule
\multicolumn{7}{c}{\textit{Phase 2 Results (MOEA/D)}} \\
\midrule
\# Phase 2 Pareto Solutions                      & 7.30    & 8.00  & 4.70  & 8.10  & 4.70  & 8.50 \\
\# Phase 2 Pareto Solutions Dominating Phase 1                     & 2.90    & 3.30  & 2.00  & 3.50  & 2.40  & 3.90 \\
Final HV (Phase 1 + Phase 2)   & 0.861   & 0.870 & 0.853 & 0.865 & 0.863 & 0.883 \\
\bottomrule
\end{tabular}%
}
\end{table*}

\begin{table*}[htbp]
\centering
\scriptsize
\caption{Pruning results for CIFAR-100. Baseline model size and accuracy, Phase 1 pruning outcomes (best, heavy, light anchor solutions), and Phase 2 multi-objective results (number of local and non-dominated Pareto solutions and the combined hypervolume). Results averaged over 10 runs. }
\label{tab:merged_comparison_C100}
\resizebox{\textwidth}{!}{%
\begin{tabular}{lcccccc}
\toprule
\textbf{Metric} & \textbf{ResNet18} & \textbf{ResNet50} & \textbf{ResNet56} & \textbf{ResNet101} & \textbf{ResNet110} & \textbf{ResNet152} \\
\midrule
\multicolumn{7}{c}{\textit{Baseline (M)}} \\
\midrule
Baseline Weights (M)  & 11.23    & 23.71  & 0.862    & 42.70  & 1.731  & 58.35 \\
Baseline Accuracy (\%)            & 76.26    & 77.92 & 68.12 & 79.24 & 63.38 & 79.44 \\
\midrule
\multicolumn{7}{c}{\textit{Phase 1 Results}} \\
\midrule
Best Solution-Non-Zero Weights (M)  & 7.810   & 8.145  & 0.819  & 18.06  & 1.764  & 24.32 \\
Best Solution Accuracy (\%)         & 76.11    & 77.31 & 67.94 & 79.08 & 63.02 & 78.98 \\
Heavy Anchor Weights (M)            & 7.528    & 6.172  & 0.513  & 15.21  & 1.510  & 22.16 \\
Heavy Accuracy (\%)                 & 75.95    & 77.19 & 67.72 & 78.85 & 62.81 & 78.02 \\
Light Anchor Weights (M)            & 6.431    & 3.912  & 0.342  & 13.10  & 1.185  & 18.84 \\
Light Accuracy (\%)                 & 73.51    & 75.25 & 65.71 & 76.17 & 61.29 & 75.94 \\
Difference (H–L, M)                 & 1.097    & 4.34  & 0.096 & 2.190  & 0.437  & 3.992 \\
Phase 1 HV                          & 0.761    & 0.764 & 0.714 & 0.759 & 0.695 & 0.772 \\
\midrule
\multicolumn{7}{c}{\textit{Phase 2 Results (NSGA-II)}} \\
\midrule
\# Phase 2 Pareto Solutions           & 6.00    & 6.40  & 4.00  & 7.10  & 4.00  & 7.20 \\
\# Phase 2 Pareto Solutions Dominating Phase 1       & 2.20   & 2.30  & 1.50  & 3.60  & 1.80  & 3.60 \\
Final HV (Phase 1 + Phase 2)             & 0.814    & 0.818 & 0.742 & 0.785 & 0.727 & 0.805 \\
\midrule
\multicolumn{7}{c}{\textit{Phase 2 Results (MOEA/D)}} \\
\midrule
\#Phase 2 Pareto Solutions            & 6.10    & 6.30  & 4.20  & 7.10  & 4.10  & 7.40 \\
\# Phase 2 Pareto Solutions Dominating Phase 1         & 2.20    & 2.20  & 1.70  & 3.50  & 2.00  & 3.80 \\
Final HV (Phase 1 + Phase 2)              & 0.817   & 0.811 & 0.749 & 0.783 & 0.731 & 0.817 \\
\bottomrule
\end{tabular}%
}
\end{table*}

Phase~1 produces the initial accuracy-sparsity trade-off through continuous threshold-based pruning. The results show that substantial parameter reductions can be achieved with minimal accuracy degradation. For example, ResNet18 is reduced from 11.18M to 6.97M parameters (a 37.6\% reduction) while retaining 93.14\% accuracy, and ResNet152 is reduced from 58.16M to 22.62M parameters (a 61.1\% reduction) with only a modest decline in accuracy. Across architectures, Phase~1 identifies a well-structured Pareto front with HV values in the range of 0.806--0.824, indicating a relatively complete initial frontier. Importantly, Phase~1 also provides two anchor points: a heavy solution with high accuracy and a light solution with aggressive pruning, that define the search interval subsequently explored in Phase~2. Deeper networks such as ResNet50, ResNet101, and ResNet152 exhibit wider heavy-to-light parameter spans, reflecting their richer pruning capacity.

Phase~2 performs a fine-grained evolutionary search using binary pruning masks derived from the Phase~1 anchors. Both NSGA-II and MOEA/D successfully discover additional Pareto-efficient configurations that enhance the accuracy--sparsity trade-off beyond what is achievable with continuous pruning alone. The evolutionary algorithms typically identify between four and eight local Pareto solutions per model, with larger architectures yielding more local optima. Across architectures, MOEA/D tends to generate slightly more local Pareto solutions than NSGA-II, suggesting a marginally stronger exploration capability within the restricted anchor-defined search interval.

However, only a subset of these local candidates becomes globally non-dominated after merging with the Phase~1 front. MOEA/D consistently produces an equal or slightly higher number of globally non-dominated solutions compared to NSGA-II, indicating a modest advantage in refining the Pareto frontier. This behavior is consistent with the sharply defined nature of the accuracy--sparsity frontier for CIFAR-10 classifiers, where only a limited number of configurations can meaningfully extend the global trade-off surface.

A key indicator of the effectiveness of Phase~2 is the improvement in HV when its solutions are merged with the Phase~1 front. For every architecture, the overall HV increases noticeably. NSGA-II raises the HV by approximately 0.044--0.058, while MOEA/D generally achieves slightly larger gains. For instance, the HV for ResNet50 increases from 0.807 in Phase~1 to 0.862 with NSGA-II and to 0.870 with MOEA/D. Similarly, for ResNet152, the HV improves from 0.821 to 0.879 and 0.883, respectively. These consistent HV improvements demonstrate that Phase~2 contributes genuinely superior trade-off solutions that were not accessible through continuous thresholding alone. Rather than merely reconfiguring the existing front, the evolutionary refinement strictly strengthens it by discovering new non-dominated points that extend the frontier toward higher sparsity at comparable accuracy or toward higher accuracy under similar parameter budgets.

\smallskip
\noindent\textbf{CIFAR-100}.Table~\ref{tab:merged_comparison_C100} summarizes the pruning results for the CIFAR-100 dataset, averaged over ten independent runs, while the resulting accuracy–sparsity trade-offs are visualized in Fig.~\ref{fig:PF_combined}. As expected, CIFAR-100 poses a substantially more difficult classification problem than CIFAR-10, and this is reflected in the lower baseline accuracies across all architectures. Even so, the preprocessing step removes redundant parameters without compromising baseline performance, preparing each model for the two-phase pruning process.

Phase~1 establishes an initial Pareto front that already exhibits significant compressibility across all ResNet variants. The best Phase~1 solutions preserve accuracies close to their baselines while achieving large reductions in non-zero parameters, for example, ResNet50 compresses from 23.71M to 8.15M parameters with accuracy maintained at 77.31\%, and ResNet152 reduces from 58.35M to 24.32M with only a minor drop in accuracy. The heavy and light anchor models form a well-defined interval in parameter space from which Phase~2 begins its evolutionary search. The Phase~1 HV values, which range from 0.695 to 0.772, reflect the increased difficulty of CIFAR-100 but still provide strong starting trade-offs.




For Phase~2 on CIFAR-100, both NSGA-II and MOEA/D consistently discover new non-dominated solutions beyond those obtained in Phase~1, confirming the benefit of discrete structural refinement. Across architectures, the two algorithms identify a comparable number of local Pareto solutions (typically between 4 and 7), with MOEA/D showing a slight tendency to produce more candidates in deeper networks. Similarly, the number of Phase~2 solutions dominating Phase~1 is closely matched, although MOEA/D provides marginally higher counts in several larger models, indicating slightly stronger refinement capability.

The improvement is also reflected in HV after merging Phase~2 with Phase~1. All architectures exhibit consistent HV gains. For example, in ResNet101 the HV increases to 0.785 with NSGA-II and to 0.783 with MOEA/D, while in ResNet152 it reaches 0.805 and 0.817, respectively. Although the differences between the two optimizers remain modest, MOEA/D generally attains slightly higher final HV values in the deeper models, suggesting improved coverage of the accuracy–sparsity trade-off surface.

Overall, the CIFAR-100 results reinforce the generality of the proposed two-phase pruning strategy. Phase~1 establishes strong continuous fronts, and Phase~2 reliably strengthens them through targeted binary mask exploration. While both NSGA-II and MOEA/D are effective, MOEA/D demonstrates a small but consistent advantage in final HV on the more complex architectures, indicating slightly better refinement performance in higher-dimensional search spaces.

\subsection{Pareto front Analysis}
\begin{figure*}
    \centering
    \begin{tabular}{cc}
        \includegraphics[width=0.48\textwidth]{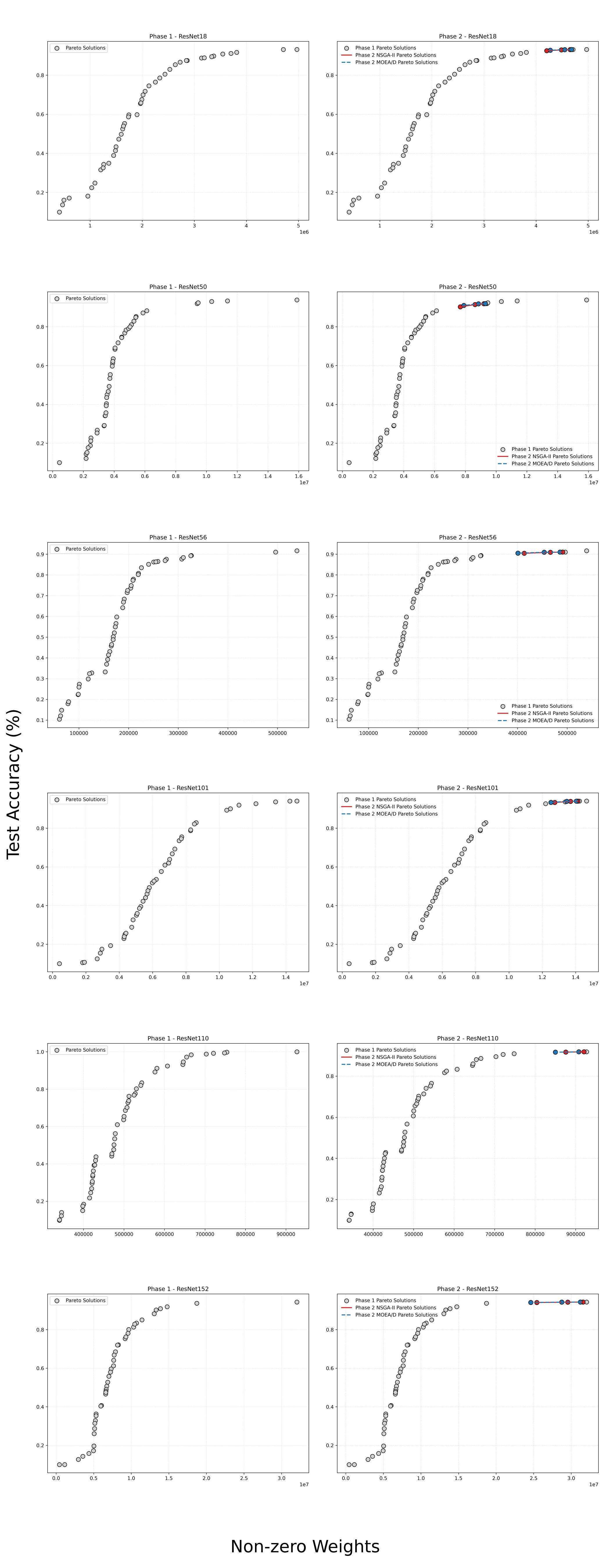} &
        \includegraphics[width=0.48\textwidth]{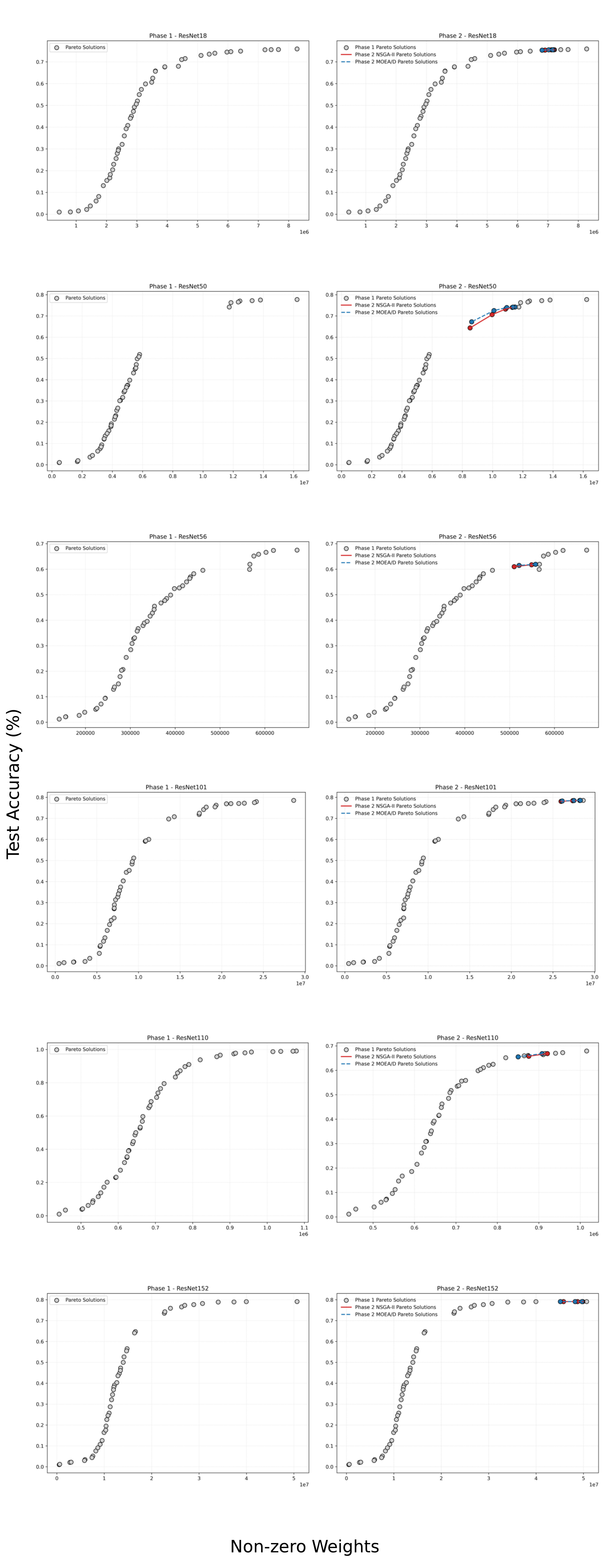} \\
        (a) CIFAR-10 & (b) CIFAR-100
    \end{tabular}
    \caption{Pareto fronts for all ResNet architectures on CIFAR-10 (a) and CIFAR-100 (b). 
    In each dataset, the left column shows Phase 1 solutions, while the right column shows 
    Phase 2 solutions generated by the multi-objective evolutionary search (NSGA-II / MOEA/D). 
    Each plot illustrates the accuracy–sparsity trade-off, highlighting how Phase 2 expands 
    and refines the Pareto front beyond Phase 1.}
    \label{fig:PF_combined}
\end{figure*}

The qualitative behavior of the combined Pareto fronts, visualized in Fig.~\ref{fig:PF_combined}, confirms this observation. Phase~2 solutions expand the frontier in two important directions: they push the boundary toward significantly higher sparsity while preserving accuracy, and they tighten the frontier around the knee region by uncovering more precise structural configurations. NSGA-II tends to enhance solution diversity near the central trade-off region, whereas MOEA/D often produces more extreme sparsity-driven solutions. Together, these behaviors explain the robust increases in hypervolume and demonstrate the complementarity of continuous pruning and evolutionary refinement.

Overall, the proposed two-phase framework achieves 35--70\% model compression with minimal accuracy loss and generates markedly superior Pareto fronts compared to relying on continuous pruning alone. The consistent improvements in hypervolume, the emergence of new non-dominated solutions, and the clear visual expansions of the fronts confirm the effectiveness of combining importance-guided initialization with binary evolutionary pruning on CIFAR-10.


\subsection{Comparison with State-of-the-Art Pruning Methods}

To begin, the primary objective of our method is to produce a denser Pareto frontier, enabling a finer-grained trade-off between compression ratio and accuracy, there is currently no existing pruning approach that explicitly optimizes for the Pareto frontier density as a standalone objective. Consequently, a direct comparison along this dimension is not feasible. For this reason, we compare our approach against the best reported solutions from state-of-the-art pruning methods in terms of accuracy–compression trade-offs.

It is important to note that absolute baseline error rates reported across different pruning methods are not directly comparable, as they depend on method-specific training protocols, initialization strategies, and fine-tuning procedures. This variability is common in pruning benchmarks, and some methods do not report baseline performance or employ architecture-specific heuristics. Consequently, the evaluation of pruning effectiveness using the relative accuracy degradation (difference between pre- and post-pruning error) together with the pruning ratio, which provides a method-agnostic measure of how much model capacity can be removed before performance degrades. This evaluation reflects the core objective of pruning: maximizing compression while minimizing accuracy loss, independent of absolute baseline accuracy.

\noindent\textbf{CIFAR-10.}  To evaluate the effectiveness of the proposed two-phase pruning framework, it is compared against several state-of-the-art evolutionary and structured pruning methods using ResNet56 and ResNet110 on the CIFAR-10 dataset. All methods share similar experimental settings in terms of dataset, baseline architectures, and fine-tuning procedures to ensure a fair comparison. The results, summarized in Table~\ref{tab:pruning_results}, report baseline and post-pruning error rates, the corresponding differences, and overall pruning rate. Some entries are marked as ``--'' because the corresponding baseline or post-pruning metrics were not reported in the original publications. In addition, not all pruning methods evaluate their approach on both CIFAR-10 and CIFAR-100 or on the same network depths, resulting in a different number of competitors across the two tables. Wherever possible, we report results exactly as provided in the original works and restrict comparisons to settings that are directly reported to avoid introducing assumptions or reimplementations.

\begin{table*}[ht]
\centering
\scriptsize
\caption{Comparison of pruning on the CIFAR-10 dataset across several evolutionary and structured pruning methods.  \textit{Baseline Error} represents the model’s error before pruning, \textit{Error after pruning} is the error following pruning, \textit{Difference} indicates the change in error, and \textit{Pruned} shows the proportion of parameters removed.}
\label{tab:pruning_results}
\begin{tabular}{llcccc}
\toprule
\textbf{Method} & \textbf{CNN architecture} & \textbf{Baseline Error (\%)} & \textbf{Error after pruning (\%)} & \textbf{Difference (\%)} & \textbf{Pruned (\%)} \\
\midrule

\multirow{2}{*}{Greedy-Filter~\cite{Li2016PruningFilters}} 
& ResNet56  & 6.96 & 6.94 & -0.02 & 27.67 \\
& ResNet110 & 6.47 & 6.70 & -0.23 & 38.63 \\
\midrule


\multirow{2}{*}{Auto-Balanced Filter~\cite{Ding2018AutoBalanced}} 
& ResNet56  & 6.96 & 8.56 & 1.60 & 41.51 \\
& ResNet110 & -- & 8.27 & -- & 40.00 \\
\midrule

\multirow{2}{*}{DeepPruningES~\cite{Fernandes2020ESPruning}} 
& ResNet56  & 6.63 & 8.12 & 1.49 & 21.30 \\
& ResNet110 & 6.20 & 7.42 & 1.22 & 16.74 \\
\midrule

\multirow{2}{*}{WIEA~\cite{Chung2024WIEA}} 
& ResNet56  & 6.72 & 7.82 & 1.10 & 20.84 \\
& ResNet110 & 7.47 & 7.69 & 0.22 & 46.23 \\
\midrule

\multirow{2}{*}{Bi-CNN-Pruning~\cite{Louati2024BiLevelPruning}} 
& ResNet56  & 6.90 & 5.77 & -1.13 & 23.21 \\
& ResNet110 & 6.20 & 6.41 & -0.21 & 26.33 \\
\midrule

\multirow{2}{*}{\textbf{Two-Phase Pruning}} 
& \textbf{ResNet56}  & \textbf{7.33} & \textbf{8.56} & \textbf{1.23} & \textbf{50.90} \\
& \textbf{ResNet110} & \textbf{6.82} & \textbf{7.89}  & \textbf{1.07} & \textbf{40.89} \\
\bottomrule
\end{tabular}
\end{table*}

The proposed Two-Phase Multi-Objective Pruning framework achieves competitive compression while maintaining comparable classification performance. Specifically, ResNet56 achieves a pruning ratio of 50.9\%, higher than any other reported method, with only a 1.23\% increase in error, and ResNet110 achieves 40.89\% pruning, again higher than other reported methods, with a 1.07\% increase in error. These results demonstrate that the two-phase optimization can substantially reduce network complexity while preserving accuracy within acceptable bounds, outperforming several prior methods in terms of achieved sparsity at similar or lower accuracy degradation.

\smallskip
\noindent\textbf{CIFAR-100.} To further examine the generalization ability of the proposed two-phase pruning framework, it is evaluated on the CIFAR-100 dataset, again using ResNet56 and ResNet110. As with the CIFAR-10 experiments, all comparative methods are run under consistent settings, including identical model architectures, training protocols, and fine-tuning procedures, ensuring a fair and controlled comparison. Table~\ref{tab:pruning_results_C100} summarizes the results, reporting baseline and post-pruning error rates, the resulting differences, and the proportion of parameters removed.

\begin{table*}[ht]
\centering
\scriptsize
\caption{Comparison of Pruning results obtained on CIFAR100 Dataset.}
\label{tab:pruning_results_C100}
\begin{tabular}{llcccc}
\toprule
\textbf{Method} & \textbf{CNN architecture} & \textbf{Baseline Error (\%)} & \textbf{Error after pruning (\%)} & \textbf{Difference (\%)} & \textbf{Pruned (\%)} \\
\midrule


\multirow{2}{*}{DeepPruningES~\cite{Fernandes2020ESPruning}} 
& ResNet56  & 41.94 & 40.20 & -1.74 & 16.18 \\
& ResNet110 & 50.17 & 50.98 & 0.81 & 17.78 \\
\midrule

\multirow{2}{*}{Bi-CNN-Pruning~\cite{Louati2024BiLevelPruning}} 
& ResNet56  & 41.86 & 24.83 & -17.03 & 16.38 \\
& ResNet110 & 50.47 & 38.44 & -12.03 & 18.13 \\
\midrule

\multirow{2}{*}{\textbf{Two-Phase Pruning}} 
& \textbf{ResNet56}  & \textbf{31.88} & \textbf{32.84} & \textbf{0.86} & \textbf{22.13} \\
& \textbf{ResNet110} & \textbf{36.62} & \textbf{37.89}  & \textbf{1.27} & \textbf{20.62} \\
\bottomrule
\end{tabular}%

\end{table*}
The results demonstrate that the proposed Two-Phase Multi-Objective Pruning framework achieves competitive performance while maintaining higher pruning ratios than most evolutionary baselines. On ResNet56, the method prunes 22.13\% of parameters with only a 0.86\% increase in error, and on ResNet110 it removes 20.62\% of parameters with just a 1.27\% degradation. Notably, the accuracy drop remains within a narrow and predictable range despite the increased complexity of CIFAR-100.


Compared to existing methods, the proposed two-phase framework explicitly prioritizes higher pruning ratios while constraining accuracy degradation to remain within a narrow and stable range. Across both ResNet56 and ResNet110, the method consistently removes a larger proportion of network parameters than most evolutionary and structured baselines, indicating a more aggressive yet controlled compression strategy. In contrast, competing approaches that exhibit smaller accuracy loss or even gains typically do so at substantially lower pruning ratios, reflecting more conservative pruning behavior. By producing a Pareto set of solutions rather than a single operating point, the proposed framework further enables selection of models based on desired pruning ratios under bounded accuracy loss, without relying on architecture-specific heuristics.

\section{Discussion}

The results across CIFAR-10 and CIFAR-100 demonstrate that the proposed two-phase pruning framework is highly effective at enabling large-scale optimization for deep neural network compression. By combining a global continuous pruning stage with a local evolutionary refinement stage, the method transforms a high-dimensional pruning problem into a tractable optimization pipeline. Phase~1 performs global threshold-based pruning and identifies two strong anchor models that define a compact search interval. This reduces the complexity of the pruning landscape and ensures that Phase~2 focuses on the most promising accuracy--sparsity region.

Phase~2 then performs a fine-grained evolutionary search using binary masks initialized through importance-guided sampling. Both NSGA-II and MOEA/D successfully discover new sparse Pareto-optimal solutions that continuous pruning alone cannot reach. Across all architectures, the merged Pareto fronts show clear expansions toward higher sparsity at comparable accuracies, as well as toward higher accuracies at similar parameter budgets. The consistent increases in hypervolume (typically +0.04 to +0.06 for NSGA-II and slightly higher for MOEA/D) confirm that the local evolutionary refinement contributes genuinely superior trade-offs to the global front. In effect, the two-phase design is successful in exposing new regions of the Pareto surface and improving frontier density.

Overall, the proposed framework achieves competitive compression performance compared to existing pruning approaches in the literature. Compression rates of 35--70\% are obtained with minimal accuracy loss, and the resulting Pareto fronts are well-structured, diverse, and robust across architectures. The combined global–local design therefore provides an effective strategy for exploring sparse solution spaces that would otherwise be inaccessible to purely continuous or purely evolutionary methods.


A primary limitation of the current framework is its reliance on unstructured weight pruning. While unstructured sparsity provides fine-grained control and enables the evolutionary algorithm to explore a rich configuration space, it does not directly translate into practical speedups on most hardware platforms. Sparse inference typically requires specialized kernels or hardware support that is not widely available on edge devices, making deployment difficult. Additionally, the binary encoding used in Phase~2 becomes large for deep architectures, increasing memory use and computational cost during evolution. Finally, unstructured binary masks are less interpretable than structured pruning decisions such as channel or block removal.

\section{Conclusion}

This work introduced an importance-guided two-Phase multi-objective pruning framework that makes large-scale CNN pruning tractable by combining global continuous pruning with local binary evolutionary refinement. Phase~1 performs broad exploration and reduces dimensionality of the optimization problem through threshold-based pruning, while Phase~2 focuses on fine-grained accuracy–sparsity trade-offs using importance-aware binary masks.
Across experiments on six popular CNN architectures and the CIFAR-10 and CIFAR-100 benchmark datasets, the method consistently discovers new Pareto-optimal solutions increasing hypervolume of the final Pareto front. These results confirm that the global–local design is effective at exploring large pruning spaces and remains competitive with existing approaches. Notably, the framework effectively handles extremely high-dimensional binary search spaces, where conventional evolutionary pruning methods struggle due to combinatorial explosion. By progressively shrinking the search space and guiding refinement using importance scores, it enables aggressive sparsity levels without sacrificing predictive performance.
Future work will extend this approach toward structured and hardware-aware pruning by operating at the level of channels and filters, enabling improved interpretability and practical acceleration on real hardware. 

\bibliographystyle{elsarticle-harv}
\bibliography{references}

\end{document}